\documentclass[10pt,twocolumn,letterpaper]{article}
\usepackage{cvpr}
\usepackage{times}
\usepackage{epsfig}
\usepackage{graphicx}
\usepackage{amsmath}
\usepackage{amssymb}
\usepackage{helvet}
\usepackage{courier}
\usepackage{upgreek}
\usepackage{color}
\usepackage{url}
\usepackage{cite}
\usepackage{booktabs}
\usepackage{multirow}
\usepackage{diagbox}
\usepackage{bm}
\usepackage{cases}
\usepackage{colortbl}
\usepackage{authblk}

\usepackage[breaklinks=true,bookmarks=false]{hyperref}
\cvprfinalcopy % *** Uncomment this line for the final submission
\def\cvprPaperID{562}

\ifcvprfinal\fi
\begin{document}

\title{Associate-3Ddet: Perceptual-to-Conceptual Association for 3D Point Cloud Object Detection}

\author[1]{Liang Du$\dagger$\thanks{duliang@mail.ustc.edu.cn. \\\hspace*{0.22cm} $\dagger$ The first two authors contributed equally to this work.}}
\author[2]{Xiaoqing Ye$\dagger$}
\author[2]{Xiao Tan}
\author[1]{Jianfeng Feng}
\author[3]{Zhenbo Xu}
\author[2]{Errui Ding}
\author[2]{Shilei Wen}

\affil[1]{Institute of Science and Technology for Brain-Inspired Intelligence, Fudan University, China, Key Laboratory of Computational Neuroscience and Brain-Inspired Intelligence (Fudan University), Ministry of Education, China. }
\affil[2]{Baidu Inc., China. }
\affil[3]{University of Science and Technology of China, China. }
\renewcommand\Authands{ and }

\maketitle
\thispagestyle{empty}

\begin{abstract}
Object detection from 3D point clouds remains a challenging task, though recent studies pushed the envelope with the deep learning techniques. Owing to the severe spatial occlusion and inherent variance of point density with the distance to sensors, appearance of a same object varies a lot in point cloud data. Designing robust feature representation against such appearance changes is hence the key issue in a 3D object detection method. In this paper, we innovatively propose a domain adaptation like approach to enhance the robustness of the feature representation. More specifically, we bridge the gap between the perceptual domain where the feature comes from a real scene and the conceptual domain where the feature is extracted from an augmented scene consisting of non-occlusion point cloud rich of detailed information. This domain adaptation approach mimics the functionality of the human brain when proceeding object perception. Extensive experiments demonstrate that our simple yet effective approach fundamentally boosts the performance of 3D point cloud object detection and achieves the state-of-the-art results.
\end{abstract}

\section{Introduction}
\label{sec:intro}
3D object detection \cite{luo2018fast, du2018general, liang2018deep, yu2018multi, yang2019std} received widespread attention from both industry and academia due to its crucial role in autonomous driving \cite{geiger2012we}.
Despite the tremendous success achieved in recent years in object detection from 2D images \cite{liu2016ssd, ren2015faster, redmon2016you, lin2017feature, duan2019centernet}, object detection based on 3D point clouds remains an open and highly challenging problem due to the object occlusion and the variance of the point distribution.

\begin{figure}[t]
 \begin{center}
  \includegraphics[width=7.4cm]{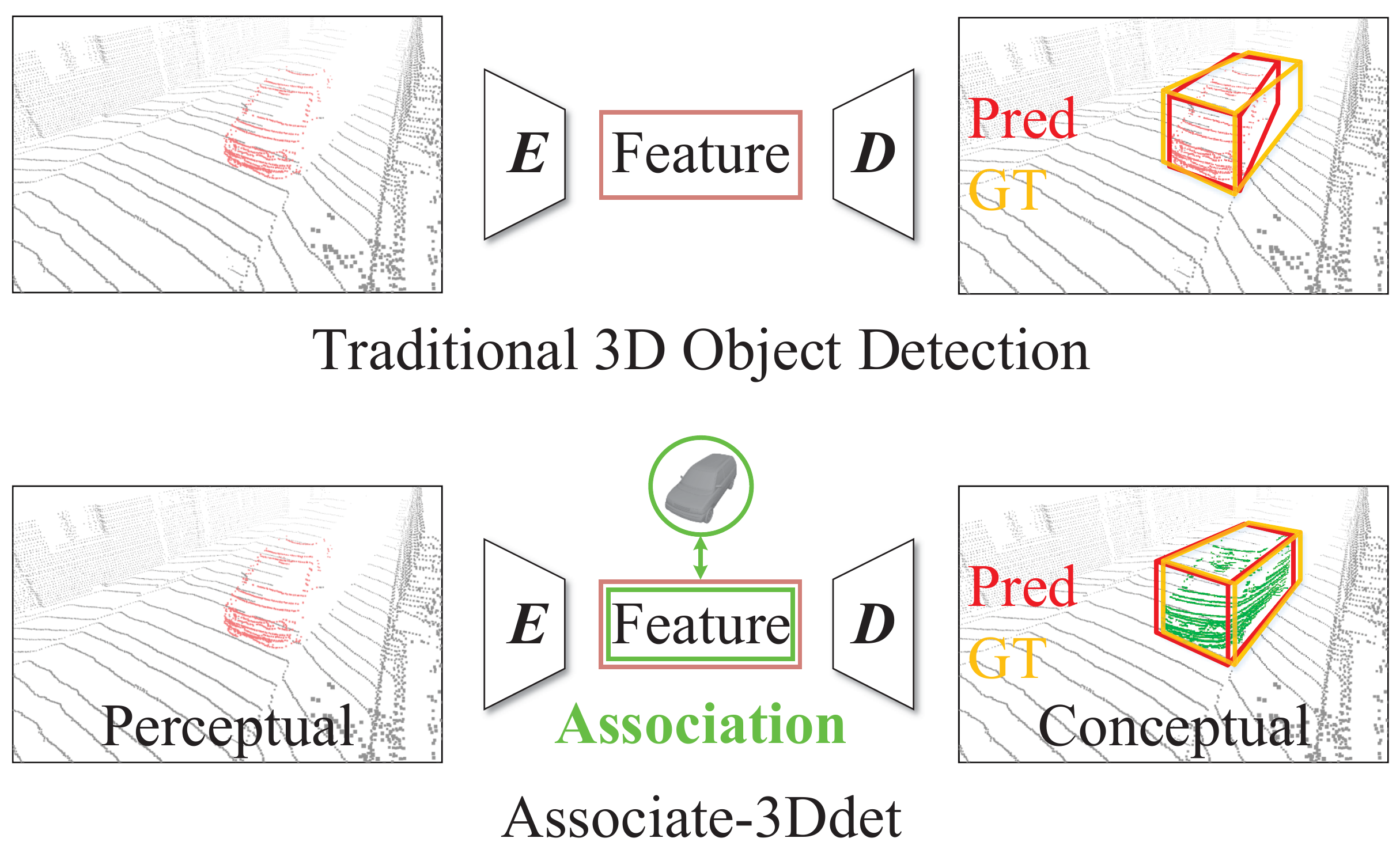}
 \end{center}
 \caption{Comparison between traditional 3D object detection and our Associate-3Ddet. $E$ and $D$ are the encoder and decoder, respectively. In contrast to the traditional method that directly uses the features of sparse point clouds for detection, our Associate-3Ddet learns to associate incomplete percepted features of objects with more complete features of corresponding class-wise conceptual models for feature enhancement.}
 \label{fig0}
\end{figure}

With the distance to the LiDAR sensor increases, the density of the point cloud dramatically decreases, resulting in the huge density variance. Moreover, some parts of objects might be invisible due to occlusion or low density point cloud. In these cases, 3D detection results are error-prone due to lack of compact perceptual features. Recent 3D detection methods \cite{yang2019std, wang2019range} struggle to alleviate this problem. On one hand, STD \cite{yang2019std} proposes PointsPool to transform intermediate point features from sparse expression to more compact voxel representation for final box prediction. However, for far objects, simple voxelization did not help the network learn compact perceptual features as most voxels might be empty. On the other hand, RANGE \cite{wang2019range} exploits generative adversarial networks to encourage consistent features for far-range as near-range objects. However, it focuses on range domain adaptation and ignores object-wise constrains like viewing angles and 3D shape. By comparison, we select objects with similar rotation angles and similar shapes as the conceptual guidance and directly adapt object-wise features from the perceptual domain to the conceptual domain following a transfer learning paradigm (see Figure \ref{fig0}).

From a psychological perspective, object perceptual is usually carried out in form of an association process. Such association process helps mapping from the observed occluded or distant object to a complete object with the finest details. This process is regarded as the associative recognition in the human cognitive system. More specifically, as it is proposed in~\cite{de2000disorders, carlesimo1998associative} that human object perceptual is a hierarchical process consisting of two stages: (a) a ``viewer-centered'' representation stage, where the features of the object are presented from the viewer's perspective , while the existing features may be incomplete due to occlusion and distance; (b) an ``object-centered'' representation stage, where the object's features are associated with its class-wise conceptual model stored in the brain.

Previous works such as \cite{hassabis2017neuroscience, cheng2018dual, du20203dcfs} demonstrate that biological plausibility can be used as a guide to design intelligent systems. Considering above issues, in this paper, we present Associate-3Ddet for object detection in point cloud, as depicted in Figure \ref{fig0}. Associate-3Ddet builds up the association between the weak perceptual features, where 3D detection results are unsatisfactory due to the dramatic variance of appearance, and the robust conceptual features. To associate the perceptual features and our proposed conceptual features, we present a self-contained method to filter conceptual models with similar view-points and high density in the same dataset, and put them on the corresponding ground-truth positions for conceptual feature extraction. Compared with perceptual features, performing 3D detection on conceptual features brings significant performance gains (see Table \ref{tab3}). By narrowing the gap between the perceptual domain and our proposed conceptual domain, Associate-3Ddet surpasses state-of-the-art on the popular KITTI 3D detection benchmark (see Table \ref{tab1} and \ref{tab2}).
The main contributions of our work are summarized as follows:
\begin{itemize}
 \item We propose a 3D object detection framework that learns to associate feature extracted from the real scene with more discriminative feature from class-wise conceptual models, which enhances the feature robustness especially for occluded objects or objects in a distance.
 \item We propose a perceptual-to-conceptual module (P2C) that adapts object features from the perceptual domain to the conceptual domain balanced by an incompletion-aware reweighting map.
 \item We present a concrete method to construct the conceptual scene without appealing to external resources.
 \item We achieve state-of-the-art 3D object detection performance on KITTI benchmark.
\end{itemize}

\section{Related Work}
\label{sec:related}

\begin{figure*}[t]
 \begin{center}
  \includegraphics[width=17.0cm]{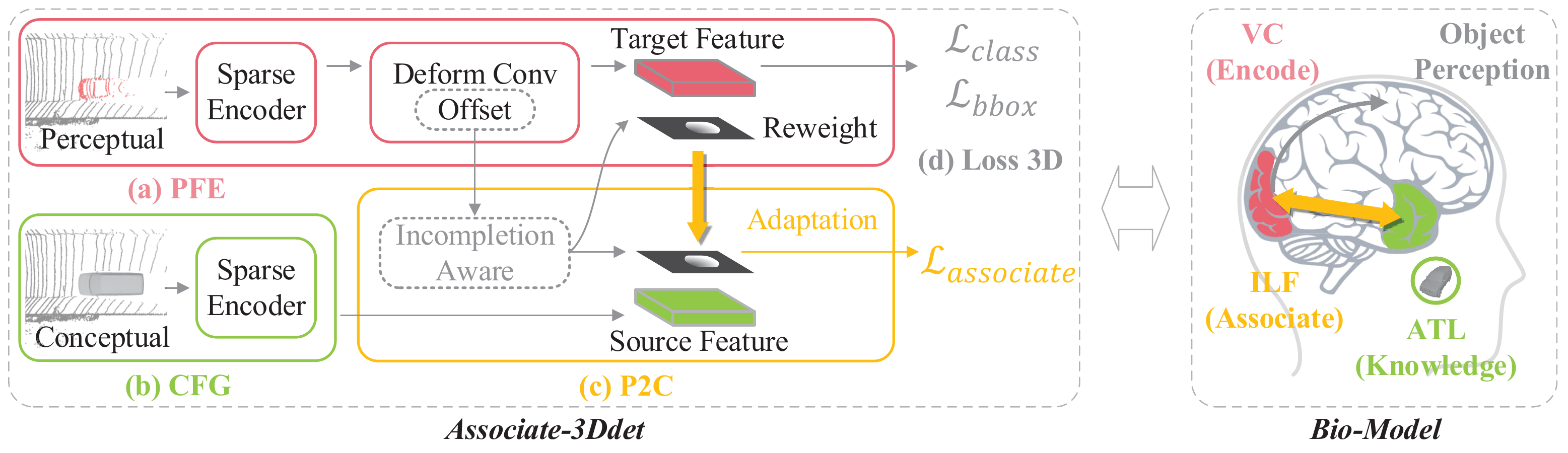}
 \end{center}
 \caption{ Overview of our proposed Associate-3Ddet framework and its underlying biological model. Associate-3Ddet mainly consists of four parts: (a) The voxel-grid representation and sparse-convolution-based perceptual feature extractor (PFE), which encodes the percepted information as the visual cortex (VC) does in the brain. (b) The conceptual feature generator (CFG) generates the features based on the constructed conceptual scene. (c) The perceptual-to-conceptual module (P2C) adapts features from the perceptual domain to the conceptual domain, which mimics the knowledge association and retrieval process in the brain. The right yellow arrow points to the adaptation loss. (d) The loss for 3D object detection. Note that the PFE without deformable convolutional layers is our baseline method. The explanation of the Bio-model can be found in Sec.\ref{2} }
 \label{fig1}
\end{figure*}

\noindent \textbf{3D Object Detection.}
Current 3D object detection methods can be divided into three categories: multi-view, voxel-based, and point-based methods.

The multi-view method \cite{chen2017multi, ku2018joint, yang2018pixor, yang2018hdnet, lang2019pointpillars} projects point clouds to the bird’s eye view (BEV) or the image plane \cite{li20173d} to perform 3D detection. \cite{chen2017multi} applied a region proposal network (RPN) to obtain positive proposals, and features are merged from the image view, front view and BEV. \cite{ku2018joint} further merged features from multiple views in the RPN phase to generate proposals.

There are several methods using voxel-grid representation. 3DFCN \cite{li20173d} and Vote3Deep \cite{engelcke2017vote3deep} discretized the point cloud on square grids and applied 3D convolutional networks to address 3D point clouds for object detection, but these approaches suffer from high computational overhead due to 3D convolutions. In \cite{zhou2018voxelnet}, the author first introduced the VoxelNet architecture to learn discriminative features from 3D point clouds for 3D object detection. VoxelNet is further improved by applying sparse convolution \cite{graham20183d, Graham2017Submanifold, yan2018second} to reduce computational consumption. Pseudo-images were utilized by \cite{lang2019pointpillars} as the representation after point clouds are voxelized, which further enhances the detection speed.

F-PointNet \cite{qi2018frustum} first exploits raw point clouds to detect 3D objects. Frustum proposals were used from off-the-shelf 2D object detectors as candidate boxes, and predictions were regressed according to interior points \cite{xu2018pointfusion}. PointRCNN \cite{shi2019pointrcnn} directly processed the raw point cloud to generate 3D proposal regions in a bottom-up manner according to the segmentation label from 3D box annotations. STD \cite{yang2019std} presented a point-based proposal generation paradigm
% for object detection on point clouds
with spherical anchors, which achieves a high recall.

Differently, RANGE \cite{wang2019range} proposed cross-range adaptation following a generative adversarial network paradigm to produce consistent features for far-range objects as near-range objects. It aims at range adaptation and ignores the object-wise features like occlusion. By comparison, our proposed object-wise domain adaptation produces better performance.\\

\noindent \textbf{Transfer Learning in Deep Neural Networks.}
Transfer learning \cite{pan2009survey} is a type of machine learning paradigm aimed at bridging different domains in natural language processing (NLP) \cite{collobert2011natural} and computer vision \cite{hoffman2014lsda}.
Our Associate-3Ddet adopts the transfer learning method to narrow the domain gap between perceptual features and class-wise conceptual features.
The key point lies in how to reduce the distribution discrepancy across different domains while avoiding over-fitting.
Deep neural networks are leveraged to perform transfer learning, since they are able to extract high-level representations which disentangle different explanatory factors of variations behind the data \cite{bengio2013representation} and manifest invariant factors underlying different populations that transfer well from source domain tasks to target domain tasks \cite{yosinski2014transferable, Du_2019_ICCV}.

\section{Methodology}
\label{sec:method}
% In this section, we introduce the proposed 3D object detection framework.
Our Associate-3Ddet establishes associations between the perceptual feature and conceptual feature based on domain adaptation.
Due to the occlusion or the long range distance, some instances only contain few points captured by LiDAR, making it difficult to directly decide the category of the object or to infer the object location knowledge. The perceptual domain corresponds to the real scene that reports unsatisfactory performance owing to the variance of appearance. Fortunately, we can construct the conceptual domain that learns from a compact point cloud or more informative instances to guide the feature learning in challenging cases.
In other words, the conceptual feature is robust to object location and orientation, which can be extracted from complete point clouds with details as fine as possible. We believe that a 3D detection algorithm will become more robust, if the gap between the two features diminishes.\par
The overall architecture of the proposed Associate-3Ddet is illustrated in Figure \ref{fig1}, which consists of four parts: (a) the perceptual feature extractor (PFE) to extract the target domain feature from the real scene; (b) the conceptual feature generator (CFG) to provide the source domain feature from an augmented scene; (c) the perceptual-to-conceptual module (P2C) to perform domain adaptation balanced by a incompletion-aware reweighting map; (d) the training loss for forcing the feature learning. % todo
We detail the network and depict the biological model that supports our motivation in Sec.\ref{2}. In Sec.\ref{3}, the training process of Associate-3Ddet is described.
The practical method we adopt to construct conceptual models is explained in Sec.\ref{4}.

\subsection{Network of Associate-3Ddet}
\label{1}
Many 3D point cloud detection methods~\cite{yan2018second, lang2019pointpillars, shi2019pointrcnn, yang2019std} can be employed as the PFE or/and the CFG branch in our approach. In this section, we instantiate one PFE and one CFG for the demonstration purpose.

\noindent \textbf{PFE to extract perceptual features.}
The PFE module functions as the feature extractor from the real scene which is shown in Figure \ref{fig1} and colored in red.
Following \cite{zhou2018voxelnet, yan2018second, lang2019pointpillars}, our Associate-3Ddet uses a brief and simple encoder that first fetches the input features by dividing the entire 3D space into voxels and then adopts sparse convolution \cite{graham20183d} for feature extraction. For sparse convolution, the output points are not computed if there is no related input point, which saves computational resources.
 % in LiDAR-based detection.

It is more difficult for the PFE branch to directly learn the conceptual feature representation from the perceptual scene comparing against the CFG branch, due to the lack of valid points within the scene.
Therefore, to make the PFE to generate feature representation resembling conceptual features for occluded objects or objects in a distance, the network may have the ability to adaptively adjust the respective field and actively capture more informative context information even in sparse cases.
Inspired by \cite{dai2017deformable} that validates the boosting capability of deformable convolution in modeling geometric transformation, this work extends 2D deformable convolutional operation to 3D voxel-based representations, as shown in Figure \ref{fig1}.

\noindent \textbf{CFG to generate conceptual features.}
A siamese sub-network named CFG extracts the conceptual feature from augmented scene serving as source domain feature.
The CFG is first separately trained end-to-end on the conceptual scene, which is derived from the real-world scene and integrated with complete object models. Here we raise that the conceptual model can be a complete point cloud for each object, such as 3D CAD models from external resources, or surrogate models with more informative knowledge originated from the same dataset, i.e., self-contained.
% We don't specifically restrain the format of the conceptual model and
One of practical conceptual model-constructing methods will be introduced in following Sec.\ref{4}.
After end-to-end training of the CFG, the parameters is fixed to provide stable feature guidance for further domain adaptation.
During the training process of the PFE, the real-world scene and its corresponding conceptual scene are fed into the PFE and CFG, respectively. Even only with sparse and partly visible point clouds of real objects, the PFE is encouraged to learn more robust features under the guidance of conceptual model. During inference process, the CFG is no longer needed.

\begin{figure}[t]
 \begin{center}
  \includegraphics[width=8.2cm]{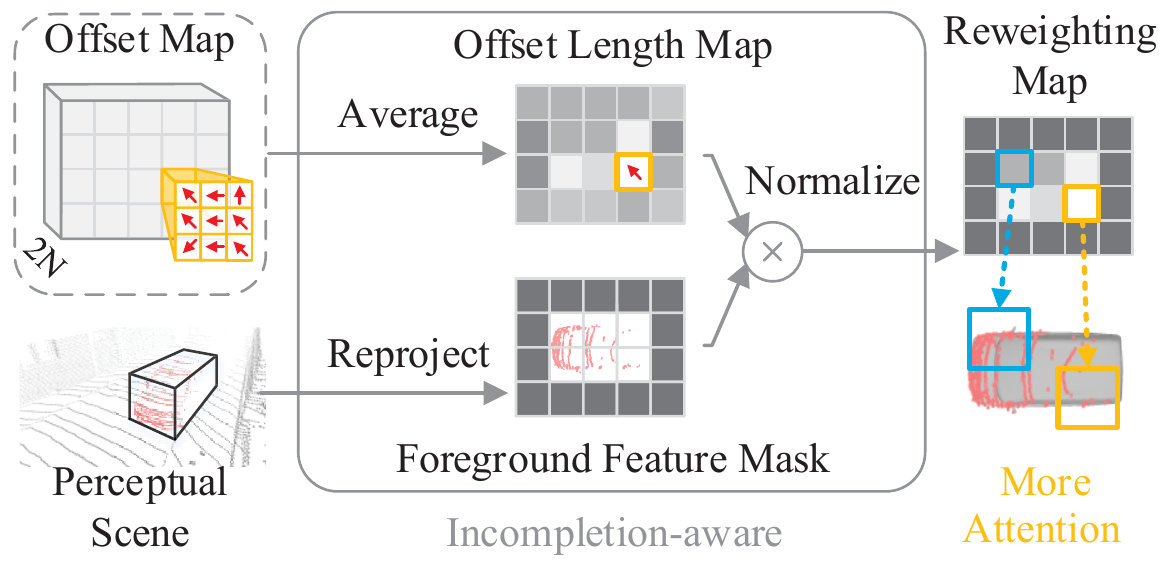}
 \end{center}
 \caption{
  Illustration of our incompletion-aware reweighting map. We calculate the average length of all learned offsets for each pixel to represent its expansion degree of searching its context information. Those regions (yellow) with incomplete and sparse object point clouds requires larger searching steps, which also needs to be given more attention during domain adaptation.
 }
 \label{fig2}
\end{figure}

\noindent \textbf{P2C to perform domain adaptation.}
The proposed P2C learns to map the perceptual feature generated by PFE to the conceptual feature extracted by the CFG to build up associations.
As mentioned above, the source domain feature is obtained by the CFG and the locations of both the real scanned objects and conceptual models are spatially aligned.
Consequently, we directly optimize the L2 distance between these two features for domain adaptation. The parameters of PFE are tuned to encourage percepted objects to generate features that resemble more informative conceptual ones. To make the domain adaptation more focused, it is restricted to foreground pixels only. As depicted in Figure \ref{fig2}, we calculate the foreground mask of the source domain by reprojecting the 3D object bounding box to the BEV view and downsample it to match the output feature map scale and the foreground pixels are those containing at least one point of objects.
% TODO
It is noticed that the pixel-wise difference between these two feature maps are prone to be large in regions containing incomplete structures due to the occlusion or sampling effect.
As proposed in \cite{dai2017deformable}, the deformable convolutional layers have the adaptive respective field according to the learned offsets.
We also observed that the learned offsets of the deformable convolutional layers in these regions are comparatively larger than those in other regions. This is probably because feature extraction for incomplete structure usually demands the surrounding information as a complement. Given such observations, we design a incompletion-aware reweighting map based on the learned offset as a guidance to reweight the per pixel L2 loss. This weighting approach shares a similar idea behind focal loss \cite{lin2017focal} and OHEM \cite{shrivastava2016training} where a weighting strategy is employed to encourage the network to focus on particular samples to achieve better performance.
As illustrated in Figure \ref{fig2}, for each pixel of the feature, the deformable convolutional layer learns $2$ ($\Delta x$ and $\Delta y$ two directions) $ \times N$ offsets. The averaged offset length over all offsets for each pixel is $\frac{1}{N} \sum_{n=1}^N\sqrt{{\Delta x}_n^2 + {\Delta y}_n^2}$, and it is first adopted as the weighting value of the pixel to obtain the offset length map. Next, the offset length map is multiplied by the foreground feature mask so that the domain adaptation is only carried out on foreground. Then, the foreground offset map is normalized to $0 \sim 1$ to get the reweighting map. The formulation will be detailed in Sec.\ref{3}. Note that after training, the whole P2C and CFG are no longer needed.

\noindent \textbf{Biological model underlying the framework.}
\label{2}
To further explain how such simple architecture is able to enhance the performance of 3D object detection, we investigate a number of biology studies on the human cognitive system corresponding to 3D object detection.
According to the biological model, the associative recognition process in the human brain involves three crucial factors, as illustrated in Figure \ref{fig1}: the visual cortex (VC), anterior temporal lobes (ATL) and inferior longitudinal fasciculus (ILF).
In general, the VC encodes the received primary information, while the ATL has been implicated as a key repository of conceptual knowledge \cite{hoffman2018percept}, which plays a critical role in object detection \cite{schacter2011psychology}. In the ATL, conceptual models of different classes are coded in an invariant manner.
The ILF is proved to be connected between the ATL and VC \cite{ashtari2012anatomy, sali2018connectomic}, which provides the associative path between conceptual models and the real-world objects.
With the enhanced object feature guided by the ATL through the ILF, the representation of an object becomes more abundant and robust for further object localization, despite of far distance and occlusion.

Our proposed P2C that adapts the object features from the perceptual domain to the conceptual domain simulates the knowledge association and retrieving process between the ATL and VC.
After perceptual and conceptual domains being well aligned after training, the network is able to adaptively generate the conceptual features without CFG. Therefore, the CFG is actually designed to build up a virtual ``repository'' of conceptual representations for the network like the ATL in the brain, and thus the CFG and P2C can be simply removed during the inference process.

\subsection{Training of Associate-3Ddet}
\label{3}
\noindent \textbf{Training of CFG.}
There are two losses for our CFG, including the binary cross entropy loss for classification and the smooth-L1 loss for 3D proposal generation.
We denote the total loss of our CFG as $\mathcal{L}_{CFG}$:
\begin{equation}
 \begin{array}{r}
  \mathcal{L}_{CFG}= \mathcal{L}_{bbox} + \mathcal{L}_{class}
 \end{array}
 \label{eq1}
\end{equation}
Same regression targets as those in \cite{yan2018second, zhou2018voxelnet} are set up and smooth-L1 loss $\mathcal{L}_{bbox}$ is adopted to regress the normalized box parameters as:
\begin{equation}
 \begin{array}{l}
  \Delta x = \frac{x_a-x_g}{d_a}, \Delta y = \frac{y_a-y_g}{h_a}, \Delta z = \frac{z_a-z_g}{d_a},    \\
  \Delta l = log(\frac{l_g}{l_a}), \Delta h = log(\frac{h_g}{h_a}), \Delta w = log(\frac{w_g}{w_a}), \\
  \Delta {\theta} = \theta_g - \theta_a,                                                             \\
 \end{array}
 \label{eq2}
\end{equation}
where $x$, $y$, and $z$ are the center coordinates; $w$, $l$, and $h$ are the width, length, and height, respectively;
$\theta$ is the yaw rotation angle; the subscripts $a$, and $g$ indicate the anchor and the ground truth, respectively;
and $d_a = \sqrt{(l_a)^2 + (w_a)^2}$ is the diagonal of the base of the anchor box.
$(x_a, y_a, z_a, h_a, w_a, l_a, \theta_a)$ are the parameters of 3D anchors and $(x_g, y_g, z_g, h_g, w_g, l_g, \theta_g)$ represent the corresponding ground truth box.

We use the focal loss introduced by \cite{lin2017focal} for classification to alleviate the sample imbalance during our anchor-based training, and the classification loss $\mathcal{L}_{class}$ is formulated as follows:
\begin{equation}
 \begin{array}{r}
  \mathcal{L}_{class}= \alpha_t(1-p_t)^\gamma log(p_t)
 \end{array}
 \label{eq3}
\end{equation}
where $p_t$ is the predicted classification probability and $\alpha$ and $\gamma$ are the parameters of the focal loss.

\noindent \textbf{Training of the whole Associate-3Ddet.}
In addition to the classification and regression loss, there is an extra loss functions for associative feature adaptation.
We denote the total loss of our Associate-3Ddet as $\mathcal{L}_{total}$:
\begin{equation}
 \begin{array}{r}
  \mathcal{L}_{total}= \mathcal{L}_{bbox} +  \mathcal{L}_{class} + \sigma \mathcal{L}_{associate}
 \end{array}
 \label{eq4}
\end{equation}
Where $\sigma$ is a hyperparameter to balance these loss terms.
The association loss $\mathcal{L}_{associate}$ for the object feature adaptation is formulated as follows:
\begin{equation}
 \begin{split}
  \mathcal{L}_{associate}= \frac{1}{P} \sum_{p=1}^P  [\Vert \mathcal{F}^p_{perceptual} - \mathcal{F}^p_{conceptual} \Vert_2 \\
  \quad \quad \quad \quad \quad \quad  \quad \quad \quad \quad \cdot (1 + \mathcal{M}^p_{reweight}) ]
  % \mathcal{L}_{associate}= \frac{1}{F} \sum\limits_{f=1}^F(\mathcal{L}_2(\mathcal{F}_{perceptual}, \mathcal{F}_{conceptual}) \cdot \mathcal{M}_{reweight})
\end{split}
 \label{eq5}
\end{equation}
where $\mathcal{F}_{perceptual}$ and $\mathcal{F}_{conceptual}$ are the two feature maps from the target and source domain. $P$ and $p$ denote the number of nonzero pixels and their indexes in $\mathcal{M}_{reweight}$, respectively. $\mathcal{M}_{reweight}$ is formulated as follows:
\begin{equation}
 \begin{array}{l}
  \mathcal{M}_{reweight} = \mathcal{\phi}(\mathcal{M}_{offset} \cdot \mathcal{M}_{foreground})
 \end{array}
 \label{eq6}
\end{equation}
where ${M}_{offset}$ denotes the average offset length map. As explained in Sec.\ref{1}, we average the length of all learned offsets for each pixel to calculate this map, and $\mathcal{M}_{foreground}$ is the reprojected foreground feature mask. $\mathcal{\phi}$ represents the operation that normalizes the map to $0 \sim 1$.

\begin{figure}[t]
 \begin{center}
  \includegraphics[width=7.0cm]{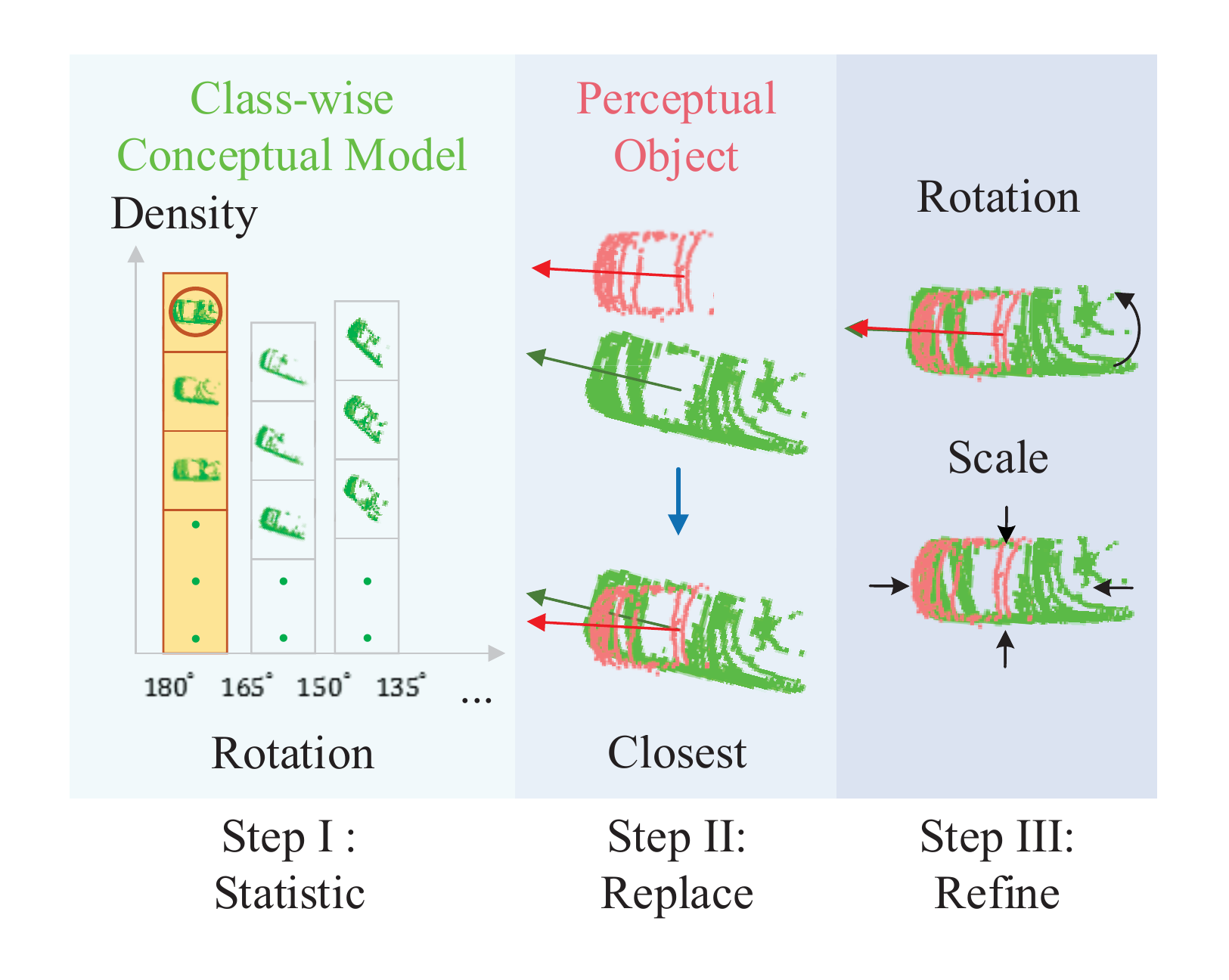}
 \end{center}
 \caption{The generation process of the conceptual scene contains three steps: (1) Objects are divided into $N$ groups according to their rotation angles. In each group, we choose the top $K\%$ objects with the most points as our conceptual models. (2) For each object not be chosen as the conceptual model, we choose the conceptual model with the minimum average closest point distance from the object as its correspondence. (3) The scale and rotation are further refined accordingly. }
 \label{fig3}
\end{figure}

\subsection{Self-contained method to build conceptual models}
\label{4}
As explained before, conceptual models can be fabricated through various approaches, such as 3D CAD models and render-based techniques. Instead of appealing to costly external resources, we present a self-constrained conceptual model-constructing method to adopt surrogate models with more informative knowledge originated from the same dataset.
Free-of-charge ground-truth point cloud instance objects with more complete structures are chosen as candidate conceptual models since 3D point cloud objects of the same class are usually of the similar scale. The main process is shown in Figure \ref{fig3} and explained as follows.\par
Firstly, for each category, relevant point cloud instances are divided into $M$ groups according to their rotation angles ranging from $-180^{\circ}$ to $+180^{\circ}$. In each equally divided rotation range, we rank the point clouds of objects in density. The top $K\%$ instances with the most complete point clouds in each range and category are chosen as our candidate conceptual models.
Next, to build pairwise correspondence for less informative objects in real-world scanned point clouds, given a less informative percepted object (which is not selected as a candidate conceptual model) and its ground truth pose of the 3D bounding box, we select conceptual models from the candidates within the same rotation range as the less informative one. To eliminate the small angle difference, the chosen candidate is further rotated into the exact pose of the real scanned object. The one with the minimum average closest point distance from percepted object is selected as the correspondence. Then, the scale (length, width and height) of the corresponding conceptual model is further fine-tuned according to the size of less informative percepted object. Each conceptual scene is finally composed by replacing the incomplete original point cloud of object with related conceptual models in 3D space.
% The pairwise perceptual and conceptual scene are then fed into PFE and CFG, respectively. Results validate that features learned by PFE under the guidance of conceptual features contribute to 3D detection performance in real percepted scenes. After end-to-end training of CFG, the parameters of CFG are fixed to provide stable feature guidance.

\begin{figure*}[t]
 \begin{center}
  \includegraphics[width=15.2cm]{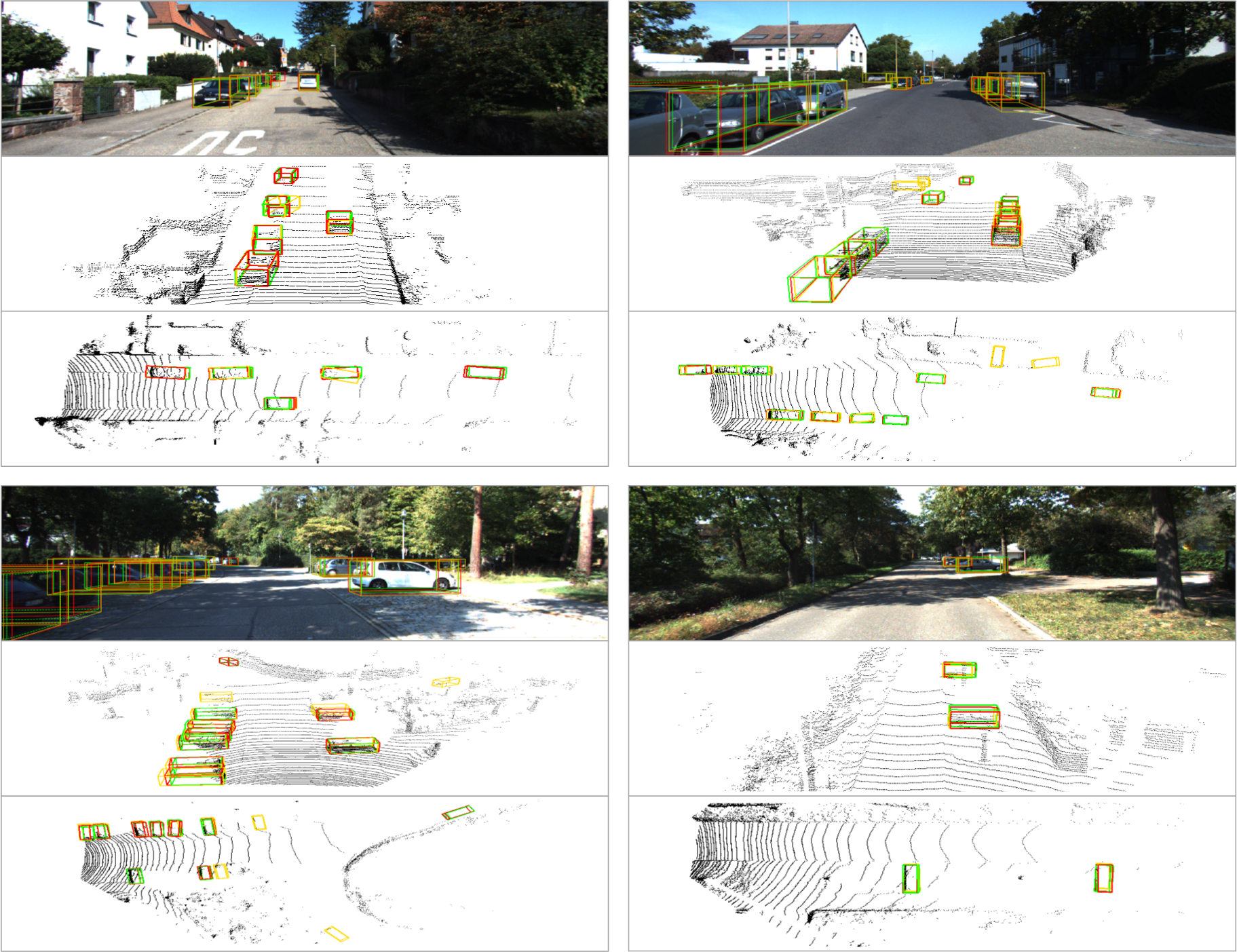}
 \end{center}
 \caption{Visualization of our results on KITTI val split set. The ground-truth 3D boxes and the predicted 3D boxes of the baseline method and our method are drawn in green, yellow and red, respectively, in the LiDAR phase. The first row shows RGB images, and the second and third rows show the front view and the bird’s-eye view, respectively.}
 \label{fig4}
\end{figure*}

\section{Experiments}
\label{sec:eval}

\begin{table*}[htb]
 \caption{Results on 3D object detection and BEV detection of the KITTI val split set at IoU = 0.7 for cars.}
 \small
 \centering
 \begin{tabular}{c|c|c|ccc|ccc|c}
  \toprule
  \hline
  \multirow{2} * {Method}              & \multirow{2} * {Time(s)} & \multirow{2} * {Modality} & \multicolumn{3}{c|}{3D Detection (\%)} & \multicolumn{3}{c|} {BEV Detection (\%)} & \multirow{2} * {GPU}                                                                  \\
                                       &                          &                           & Mod.                                   & Easy                                     & Hard                 & Mod.           & Easy           & Hard           &             \\
  \midrule
  MV3D    \cite{chen2017multi}         & 0.24                     & RGB + LiDAR               & 62.68                                  & 71.29                                    & 56.56                & 78.10          & 86.55          & 76.67          & TITAN X     \\

  AVOD-FPN     \cite{ku2018joint}      & 0.1                      & RGB + LiDAR               & 74.44                                  & 84.41                                    & 68.65                & -              & -              & -              & TITAN XP    \\
  F-PointNet    \cite{qi2018frustum}   & 0.17                     & RGB + LiDAR               & 70.92                                  & 83.76                                    & 63.65                & 84.02          & 88.16          & 76.44          & GTX 1080    \\
  \midrule
  VoxelNet \cite{zhou2018voxelnet}     & 0.22                     & LiDAR only                & 65.46                                  & 81.98                                    & 62.85                & 84.81          & 89.60          & 78.57          & TITAN X     \\
  SECOND    \cite{yan2018second}       & 0.05                     & LiDAR only                & 76.48                                  & 87.43                                    & 69.10                & 87.07          & 89.96          & 79.66          & GTX 1080Ti  \\
  RANGE   \cite{wang2019range}         & 0.05                     & LiDAR only                & 78.31                                  & 88.80                                    & 76.16                & 87.82          & 90.32          & 87.51          & GTX 1080 Ti \\
  PointRCNN    \cite{shi2019pointrcnn} & 0.1                      & LiDAR only                & 78.63                                  & 88.88                                    & 77.38                & 87.07          & 89.96          & 79.66          & TITAN XP    \\
  STD \cite{yang2019std}               & 0.08                     & LiDAR only                & 78.70                                  & 88.80                                    & \textbf{78.20}       & 88.30          & 90.10          & 87.40          & TITAN V     \\
  Ours                                 & 0.06                     & LiDAR only                & \textbf{79.17}                         & \textbf{89.29}                           & 77.76                & \textbf{88.98} & \textbf{90.55} & \textbf{87.71} & GTX 1080Ti  \\
  \bottomrule
 \end{tabular}
 \label{tab1}
\end{table*}

\subsection{Dataset and Experimental Setup}
We evaluate Associate-3Ddet on the widely acknowledged 3D object detection benchmark, KITTI dataset \cite{geiger2012we}, on `Car' category. The color images as well as the relevant point cloud are provided. We split the 7481 training images equally into train and validation set in accordance with previous works such as \cite{yang2019std}.
Average precision (AP) metrics measured in 3D and BEV are utilized to evaluate the performance of different methods. During evaluation, we follow the official KITTI evaluation protocol. Three levels of difficulty are defined according to the 2D bounding box height, occlusion and truncation degree as follows: easy, moderate and hard. The benchmark ranks algorithms based on the moderately difficult results.

\subsection{Implementation Details.}
\noindent \textbf{Data Augmentation}
We perform data augmentation to prevent over-fitting following \cite{yan2018second} on both perceptual and conceptual datasets.

\noindent \textbf{Network Architecture}
In contrast to \cite{yan2018second} that adopts stacked convolutions for input feature encoding, we simply limit the number of each voxel to be no more than five points and compute the mean x, y, z values within them as the input feature, followed by a 128-dimension linear layer. After feeding into stacked sparse and submanifold convolutions for feature extraction and dimensionality reduction, the shape of obtained encoded tensor is $128 \times 2 \times 200 \times 176$, where 2 corresponds to height dimension. We squeeze the height dimension by reshaping it into the feature map channels. Since PFE and CFG are of siamese structures with different input features, we leave the detailed architecture in the supplementary material.
Then, features of PFE are fed into a deformable convolutional layer with a 128-output feature map, a kernel size of (5, 5).
Finally, naive convolutions are applied to features originated from the deformable convolutional layer.

\noindent \textbf{Training Parameters}
Both CFG and PFE are trained with the ADAM optimizer and batch size 6 for 80 epochs on a single NVIDIA TITAN RTX card.
We utilize the cosine annealing learning rate strategy with an initial learning rate 0.001.
% The whole training process of our Associate-3Ddet networks takes about 12 hours
We set $\alpha = 0.25$ and $\gamma = 2$ in focal loss, $\sigma = 0.5$ to balance the loss terms, and $M=24$ and $K=20$ to create the conceptual scenes. We first train CFG for conceptual models end-to-end and then keep it fixed for further Associate-3Ddet network training.

\subsection{Quantitative Results}
As shown in Table \ref{tab1}, we evaluate our Associate-3Ddet on 3D detection and BEV detection benchmark on KITTI val split set.
For 3D object detection, by only utilizing LiDAR point clouds and voxel-based encoding, our proposed Associate-3Ddet outperforms most existing state-of-the-art methods and obtains comparable results on ``hard'' difficulty level with STD \cite{yang2019std}, whereas our network runs at much higher FPS (see the time cost in Table \ref{tab1} and GPU resource for detail).
For BEV detection, our method outperforms previous top-ranked methods by large margins on all the difficulty levels.
Thanks to the self-contained conceptual model and brain-inspired P2C module, our simple yet effective network is able to get superior performance to complicated networks and runs at high FPS.

Table \ref{tab2} shows the result on KITTI 3D object detection test server.
For the fairness, the comparison is carried out among methods based on pseudo-image representations.
Our Associate-3Ddet is superior to all LiDAR-based methods on all entries. Note that in this paper, we focus on pseudo-image-based methods to demonstrate the effectiveness of our brain-inspired approach. Our method can also be easily expanded to the point-based two-stage methods.

\begin{table*}[ht]
 \caption{Results on KITTI 3D object detection test server (test split). The 3D object detection and BEV detection are evaluated by average precision at IoU = 0.7 for cars. }
 \small
 \centering
 \begin{tabular}{c|c|c|ccc|ccc|c}
  \toprule
  \hline
  \multirow{2} * {Method}                  & \multirow{2} * {Time(s)} & \multirow{2} * {Modality} & \multicolumn{3}{c|}{3D Detection (\%)} & \multicolumn{3}{c|} {BEV Detection (\%)} & \multirow{2} * {GPU}                                                                 \\
                                           &                          &                           & Mod.                                   & Easy                                     & Hard                 & Mod.           & Easy           & Hard           &            \\
  \midrule

  MV3D \cite{chen2017multi}                & 0.24                     & RGB + LiDAR               & 52.73                                  & 66.77                                    & 51.31                & 77.00          & 85.82          & 68.94          & TITAN X    \\
  AVOD-FPN \cite{ku2018joint}              & 0.1                      & RGB + LiDAR               & 71.88                                  & 81.94                                    & 66.38                & 83.79          & 88.53          & 77.90          & TITAN XP   \\
  % F-PointNet \cite{qi2018frustum}          & 0.17                     & RGB + LiDAR               & 70.39                                   & 81.20                                     & 62.19                & 84.00          & 88.70          & 75.33          & GTX 1080   \\
  ContFuse \cite{liang2018deep}            & 0.06                     & RGB + LiDAR               & 66.22                                  & 82.54                                    & 64.04                & 85.83          & 88.81          & 77.33          & -          \\
  % RoarNet \cite{shin2019roarnet}           & 0.1                      & RGB + LiDAR               & 73.04                                   & 83.71                                     & 59.16                & 79.41          & 88.20          & 70.02          & TITAN X    \\
  % IPOD \cite{yang2018ipod}                 & 0.2                      & RGB + LiDAR               & 72.57                                   & 79.75                                     & 66.33                & 83.98          & 86.93          & 77.85          & Tesla P40  \\
  UberATG-MMF \cite{liang2019multi}        & 0.08                     & RGB + LiDAR               & 76.75                                  & \textbf{86.81}                           & 68.41                & 87.47          & 89.49          & 79.10          & TITAN XP   \\
  \midrule
  VoxelNet \cite{zhou2018voxelnet}         & 0.22                     & LiDAR only                & 65.11                                  & 77.47                                    & 57.73                & 79.26          & 89.35          & 77.39          & TITAN X    \\
  SECOND    \cite{yan2018second}           & 0.05                     & LiDAR only                & 73.66                                  & 83.13                                    & 66.20                & 79.37          & 88.07          & 77.95          & GTX 1080Ti \\
  PointPillars \cite{lang2019pointpillars} & 0.016                    & LiDAR only                & 74.99                                  & 79.05                                    & 68.30                & 86.10          & 88.35          & 79.83          & GTX 1080Ti \\
  % Fast Point R-CNN    \cite{chen2019fast}  & 0.065                    & LiDAR only                & 75.73                                   & 84.28                                     & 67.39                & 86.01          & 88.03          & 78.17          & Tesla P40  \\
  % PointRCNN    \cite{shi2019pointrcnn}     & 0.1                      & LiDAR only                & 75.76                                   & 85.94                                     & 68.32                & 85.68          & 89.47          & 79.10          & TITAN XP   \\
  Ours                                     & 0.06                     & LiDAR only                & \textbf{77.40}                         & 85.99                                    & \textbf{70.53}       & \textbf{88.09} & \textbf{91.40} & \textbf{82.96} & GTX 1080Ti \\
  % \hline
  \bottomrule
 \end{tabular}
 \label{tab2}
\end{table*}

\begin{table}[htb]
 \caption{The upper bound performance of CFG training with the conceptual data on $\rm AP_{3d}$.}
 \small
 \centering
 \begin{tabular}{c|c|ccc}
  \toprule
  \hline
  \multirow{2} * {training data} & \multirow{2} * {validation data} & \multicolumn{3}{c}{$\rm AP_{3d}$ (IoU=0.7)}                 \\
                                 &                                  & Mod.                                        & Easy  & Hard  \\
  \hline
  \multirow{2} * {conceptual}    & real                             & 59.87                                       & 78.58 & 59.33 \\
                                 & conceptual                       & 90.69                                       & 98.18 & 90.70 \\
  \hline
  real + conceptual              & real
                                 & 76.75                            & 87.27                                       & 74.68         \\
  \bottomrule
 \end{tabular}
 \label{tab3}
\end{table}

\begin{table}[htb]
 \caption{Ablation study for our Associate-3Ddet with different settings on $\rm AP_{3d}$.}
 \small
 \centering
 \resizebox{83mm}{20mm}{
  \begin{tabular}{c|c|ccc}
   \toprule
   \hline
   {P2C}                         & \multirow{2} * {Method}    & \multicolumn{3}{c}{$\rm AP_{3d}$ (IoU=0.7)}                                   \\
   {\& CFG}                      &                            & Mod.                                        & Easy           & Hard           \\
   \hline
   \multirow{2} * {$\times$}     & Baseline                   & 76.78                                       & 87.28          & 75.46          \\
   % & Baseline (real+conceptual) & 76.75                                       & 87.27          & 74.68          \\
                                 & Baseline (with $D$)        & 76.86                                       & 87.34          & 75.70          \\
   \hline
   \multirow{5} * {$\checkmark$} & Ours (without $M$,$D$,$R$) & 78.05                                       & 88.34          & 76.75          \\
                                 & Ours (with $D$)            & 78.29                                       & 88.52          & 76.91          \\
                                 & Ours (with $M$)            & 78.46                                       & 88.74          & 77.20          \\
                                 & Ours (with $M$,$D$)        & 78.69                                       & 88.96          & 77.36          \\
   % & Ours (with $D$,$R$)            & 78.71                                       & 89.06          & 77.50          \\
                                 & Ours (full approach)       & \textbf{79.17}                              & \textbf{89.29} & \textbf{77.76} \\
   \bottomrule
  \end{tabular}
 }
 \label{tab4}
\end{table}

\begin{table}[htb]
 \caption{Different collection strategies of conceptual models for CFG on $\rm AP_{3d}$.}
 \small
 \centering
 \begin{tabular}{c|ccc}
  \toprule
  \hline
  \multirow{2} * {$K$\%} & \multicolumn{3}{c}{$\rm AP_{3d}$ (IoU=0.7)}                                   \\
                         & Mod.                                        & Easy           & Hard           \\
  \hline
  50\%                   & 78.45                                       & 88.88          & 77.15          \\
  40\%                   & 78.49                                       & 88.96          & 77.19          \\
  30\%                   & 78.59                                       & 89.03          & 77.29          \\
  20\%                   & \textbf{79.17}                              & \textbf{89.29} & \textbf{77.76} \\
  10\%                   & 78.89                                       & 88.92          & 77.73          \\
  \bottomrule
 \end{tabular}
 \label{tab6}
\end{table}

\subsection{Ablation Study}
To validate the effectiveness of different modules and settings of Associate-3Ddet, the following ablation experiments are conducted.

\noindent \textbf{The upper bound performance of CFG.}
Table \ref{tab3} demonstrates the upper bound performance of our CFG, i.e., adopting the conceptual models for both train and test. The high precision (over 90$\%$) indicates the possibility of adopting conceptual models to guide the PFE module for enhanced feature learning. The third row ``real + conceptual'' indicates merely training baseline model with KITTI train split set and the generated conceptual data without using siamese network for adaptation. The result validates that simply mixing more training data using the same network won't boost the performance.

\noindent \textbf{The strength of domain adaptation.}
We conduct an investigation by measuring the performance with or without our P2C and CFG, as shown in Table \ref{tab4}.
The first row is the baseline results trained with only KITTI train split set.
The second row shows the baseline equipped with only $D$ (deformable convolutional layers).
We find that without domain adaptation, simply adopting deformable convolutional layers has little effect on the performance of the detector.
Owing to domain adaptation (P2C and CFG), improvement is observed in Row 3.
Following rows indicate the performance of Associate-3Ddet with or without $D$, $M$ (foreground mask) and $R$ (reweighting map).
The improvements on all difficulty levels indicate that our full approach equipped with P2C and CFG learns better discriminative and robust features for 3D detection.

\noindent \textbf{Hyperparameter for constructing conceptual models.}
To evaluate the effect of hyperparameters for constructing candidate conceptual models from original object instances, we conduct experiments with different settings of $K$, where the top $K\%$ instances in dense-to-sparse order within each angle range are selected as candidates.
Smaller $K$ represents that each candidate conceptual model contains more points, which is more close to ideal models. As is revealed in Table \ref{tab6}, smaller $K$ leads to higher performance due to more complete conceptual models being selected. However, if the percentage $K$ is too small, for each angle range, too few conceptual models exist, making it less possible to find the best corresponding conceptual model with a similar structure for each percepted instance. Comparison shows that setting $K$ to 20 achieves superior performance. More experiments can be found in the supplementary material.

\subsection{Qualitative Results}
We present some representative comparison results of the baseline and our Associate-3Ddet on the val split of KITTI in Figure \ref{fig4}. It can be seen that the occluded and distant objects are well detected after adopting the proposed association mechanism, which demonstrates that the proposed brain-inspired approach adaptively generates robust features for predicting more accurate 3D bounding boxes.

\section{Conclusions}
\label{sec:conclusion}
Inspired by human associative recognition, we propose a simple yet effective 3D object detection framework that learns to associate features of precepted objects with discriminative and robust features from their conceptual models by domain adaptation. This approach explicitly bridges the gap between two domains, and enhances the robustness against appearance changes in point clouds. In addition, our approach can be easily integrated into many existing object detection methods in 3D point clouds. Experimental results on the KITTI benchmark dataset demonstrate the effectiveness and robustness of our Associate-3Ddet.
% In future, we hope that our brain-inspired Associate-3Ddet could be extended to object detection in 2D images serving as a reliable solution for generic object detection.

\section*{Acknowledgement}
This work was supported by the 111 Project (NO.B18015), the National Natural Science Foundation of China (No.91630314), the key project of Shanghai Science \& Technology (No.16JC1420402), Shanghai Municipal Science and Technology Major Project (No.2018SHZDZX01) and ZJLab.

{\small
 \bibliographystyle{ieee_fullname}
 \bibliography{egbib}
}

\end{document}